\documentclass[10pt,twocolumn,letterpaper]{article}

\usepackage{cvpr}
\usepackage{times}
\usepackage{epsfig}
\usepackage{graphicx}
\usepackage{amsmath}
\usepackage{amssymb}

\usepackage{color}
\usepackage{enumitem}
\usepackage{algorithm}
\usepackage{multirow}
\usepackage[noend]{algorithmic}
\usepackage{epstopdf}
\usepackage{amsfonts}

\newcommand{\bC}{\mathbf{C}}
\newcommand{\bD}{\mathbf{D}}

\newcommand{\bI}{\mathbf{I}}

\newcommand{\bP}{\mathbf{P}}

\newcommand{\bS}{\mathbf{S}}
\newcommand{\bT}{\mathbf{T}}

\newcommand{\bW}{\mathbf{W}}
\newcommand{\bX}{\mathbf{X}}
\newcommand{\bY}{\mathbf{Y}}

\newcommand{\real}{\mathbb{R}}

\long\def\ignorethis#1{} 
\usepackage{amsfonts}
\usepackage{amsmath}

\newsavebox{\savepar}

\def\natural{\mathbb{N}}

\def\ie{{\em i.e.,}}

\def\eg{{\em e.g.,}}

\usepackage{graphicx}
\usepackage{paralist}
\usepackage{color,soul}
\usepackage{makecell}
\usepackage{multirow, booktabs}
\usepackage{tablefootnote}
\providecommand{\hessam}[1]{}
\providecommand{\fixme}[1]{#1}

\usepackage{subfigure}
\newcommand{\norm}[1]{\|#1\|}
\usepackage[pagebackref=true,breaklinks=true,letterpaper=true,colorlinks,bookmarks=false]{hyperref}

\cvprfinalcopy 

\def\cvprPaperID{3345} 

\ifcvprfinal\fi
\begin{document}

\title{LCNN: Lookup-based Convolutional Neural Network}

\author{Hessam Bagherinezhad$^{1,2}$ \hspace{2cm} Mohammad Rastegari$^{2,3}$ \hspace{2cm} Ali Farhadi$^{1,2,3}$\vspace{3mm}\\
$^1${\normalsize University of Washington} \hspace{1cm} $^2${\normalsize XNOR.AI} \hspace{1cm} $^3${\normalsize Allen Institute for AI}\\
\texttt{\normalsize \{hessam, mohammad, ali\}@xnor.ai}
}

\maketitle

\begin{abstract}
Porting state of the art deep learning algorithms to resource constrained compute platforms (e.g. VR, AR, wearables) is extremely challenging.  We propose a fast, compact, and accurate model for convolutional neural networks that enables efficient learning and inference. We introduce LCNN, a lookup-based convolutional neural network that encodes convolutions by few lookups to a dictionary that is trained to cover the space of weights in CNNs.  Training LCNN involves jointly learning a dictionary and a small set of linear combinations. The size of the dictionary naturally traces a spectrum of trade-offs between efficiency and accuracy. Our experimental results on ImageNet challenge show that LCNN can offer $3.2\times$ speedup while achieving $55.1\%$ top-1 accuracy using AlexNet architecture. Our fastest LCNN offers $37.6\times$ speed up over AlexNet while maintaining $44.3\%$ top-1 accuracy. LCNN not only offers dramatic speed ups at inference, but it also enables efficient training. In this paper, we show the benefits of LCNN in few-shot learning and few-iteration learning, two crucial aspects of on-device training of deep learning models.

\end{abstract}

\section{Introduction}

In recent years convolutional neural networks (CNN) have played major roles in improving the state of the art across a wide range of problems in computer vision, including image classification \cite{krizhevsky2012imagenet,simonyan2014very,szegedy2015going,kaming2016residual}, object detection \cite{girshick2014rich,girshick2015fast,ren2015faster}, segmentation \cite{pinheiro2015learning,long2015fully}, etc. These models are very expensive in terms of computation and memory. For example,  AlexNet\cite{krizhevsky2012imagenet} has 61M parameters and performs 1.5B high precision operations to classify a single image. These numbers are even higher for deeper networks, \eg VGG \cite{simonyan2014very}. The computational burden of learning and inference for these models is significantly higher than what most compute platforms can afford. 

Recent advancements in virtual reality (VR by Oculus) \cite{oculus2012oculus}, augmented reality (AR by HoloLens) \cite{gottmer2015merging}, and smart wearable devices increase the demand for getting our state of the art deep learning algorithm on these portable compute platforms. Porting deep learning methods to these platforms is challenging mainly due to the gap between what these platforms can offer and what our deep learning methods require. More efficient approaches to deep neural networks is the key to this challenge. 

Recent work on efficient deep learning have focused on model compression and reducing the computational precision of operations in neural networks~\cite{chen2015compressing,han2015deep, Rastegari2016XNORNetIC}. CNNs suffer from over-parametrization~\cite{denil2013predicting} and often encode highly correlated parameters~\cite{jaderberg14speeding}, resulting in inefficient computation and memory usage\cite{denil2013predicting}. Our key insight is to leverage the correlation between the parameters and represent the space of parameters by a compact set of weight vectors, called dictionary. In this paper, we introduce LCNN, a lookup-based convolutional neural network that encodes convolutions by few lookups to a dictionary that is trained to cover the space of weights in CNNs.  Training LCNN involves jointly learning a dictionary and a small set of linear combinations. The size of the dictionary naturally traces a spectrum of trade-offs between efficiency and accuracy. Our experimental results using AlexNet on ImageNet challenge show that LCNN can offer $3.2\times$ speedup while achieving $55.1\%$ top-1 accuracy. Our fastest LCNN offers $37.6\times$ speed up over CNN while maintaining $44.3\%$ top-1 accuracy. In the ResNet-18, the most accurate LCNN offers $5\times$ speedup with $62.2\%$ accuracy and the fastest LCNN offers $29.2\times$ speedup with $51.8\%$ accuracy  

In addition, LCNN enables efficient training; almost all the work in efficient deep learning have focused on efficient inference on resource constrained platforms~\cite{Rastegari2016XNORNetIC}. Training on these platforms is even more challenging and requires addressing two major problems: \begin{inparaenum}[i.] \item \textbf{few-shot learning:} the settings of on-device training dictates that there won't be enough training examples for new categories. In fact, most training needs to be done with very few training examples; \item \textbf{few-iteration learning:} the constraints in computation and power require the training to be light and quick. This imposes hard constraints on the number of iterations in training.\end{inparaenum} LCNN offers solutions for both of these problems in deep on-device training.

Few-shot learning, the problem of learning novel categories from few examples (sometimes even one example), have been extensively studies in machine learning and computer vision\cite{fei2006one}. The topic is, however, relatively new for deep learning\cite{hariharan2016low}, where the main challenge is to avoid overfitting. The number of parameters are significantly higher than what can be learned from few examples. LCNN, by virtue of having fewer parameters to learn (only around 7\% of parameters of typical networks), offers a simple solution to this challenge. Our dictionary can be learned offline from training data where enough training examples per category exists. When facing new categories, all we need to learn is the set of sparse reconstruction weights. Our experimental evaluations show significant gain in few-shot learning; $6.3\%$ in one training example per category. 

Few-iteration learning is the problem of getting highest possible accuracy in few iterations that a resource constrained platform can offer. In a typical CNN, training often involves hundreds of thousands of iterations. This number is even higher for recent deeper architectures. LCNN offers a solution: dictionaries in LCNN are architecture agnostic and can be transferred across architectures or layers. This allows us to train a dictionary using a shallow network and transfer it to a deeper one. As before, all we need to learn are the few reconstruction weights; dictionaries don't need to be trained again. Our experimental evaluations on ImageNet challenge show that using LCNN we can train an 18-layer ResNet with a pre-trained dictionary from a 10-layer ResNet and achieve $16.2\%$ higher top-1 accuracy on $10K$ iterations.

In this paper, we  \begin{inparaenum}[1)] \item introduce LCNN; \item show state of the art efficient inference in CNNs using LCNN; \item demonstrate possibilities of training deep CNNs using as few as one example per category \item show results for few iteration learning \end{inparaenum}.

\section{Related Work}
\label{sec:related}
A wide range of methods have been proposed to address efficient training and inference in deep neural networks. Here, we briefly study these methods under the topics that are related to our approach. 

\textbf{Weight compression:} Several attempts have been made to reduce the number of parameters of deep neural networks. Most of such methods~\cite{gong2014compressing,yang2015deep,chen2015compressing,han2015deep,srinivas2015data} are based on compressing the fully connected layers, which contain most of the weights. These methods do not achieve much improvement on speed. In~\cite{Iandola2016SqueezeNetAA}, a small DNN architecture is proposed which is fully connected free and has 50x fewer parameters in compare to AlexNet~\cite{alexnet}. However, their model is slower than AlexNet. 
Recently \cite{han2015learning, han2015deep} reduced the number of parameters by pruning. 
All of these approaches update a pre-trained CNN, whereas we propose to train a compact structure that enables faster inference.  

\textbf{Low Rank Assumption:} Approximating the weights of convolutional layers with low-rank tensor expansion has been explored by ~\cite{jaderberg14speeding,denil2013predicting}. They only demonstrated speedup in the case of large convolutions. \cite{denton2014exploiting} uses SVD for tensor decomposition to reduce the computation in the lower layers on a pre-trained CNN. \cite{zhang2015efficient} minimizes the reconstruction error of the nonlinear responses in a CNN, subject to a low-rank constraint which helps to reduce the complexity of filters.
Notably, all of these methods are a post processing on the weights of a trained CNN, and none of them train a lower rank network from scratch.


\textbf{Low Precision Networks:} 
A fixed-point implementation of 8-bit integer was compared with 32-bit floating point activations in \cite{vanhoucke2011improving, hwang2014fixed}. Several network quantization methods are proposed by~\cite{gong2014compressing, anwar2015fixed, lin2015neural, lin2015neural, hubara2016quantized}. 
Most recently, binary networks has shown to achieve relatively strong result on ImageNet~\cite{Rastegari2016XNORNetIC}. They have trained a network that computes the output with mostly binary operations, except for the first and the last layer.
\cite{courbariaux2015binaryconnect} uses the real-valued version of the weights as a key reference for the binarization process. 
\cite{courbariaux2016binarynet} is an extension of \cite{courbariaux2015binaryconnect}, where both weights and activations are binarized. 
\cite{kim2016bitwise} retrains a previously trained neural network with binary weights and binary inputs. Our approach is orthogonal to this line of work. In fact, any of these methods can be applied in our model to reduce the precision.  

\textbf{Sparse convolutions:} Recently, several attempts have been made to sparsify the weights of convolutional layers \cite{liu2015sparse,wu2015quantized,wen2016learning}. \cite{liu2015sparse} shows how to reduce the redundancy in parameters of a CNN using a sparse decomposition. 
\cite{wu2015quantized} proposed a framework to simultaneously speed up the computation and reduce the storage of CNNs. 
\cite{wen2016learning}  proposed a Structured Sparsity Learning (SSL) method to regularize the structures (i.e., filters, channels, filter shapes, and layer depth) of CNNs. 
Only in \cite{wen2016learning} a sparse CNN is trained from scratch which makes it more similar to our approach. However, our method provides a rich set of dictionary that enables implementing convolution with lookup operations.

\textbf{Few-Shot Learning:} The problem of learning novel categories has been studied in \cite{thrun1996learning, azizpour2015generic, littwin2015multiverse}. Learning from few examples per category explored by \cite{hariharan2016low}. \cite{fei2006one, vinyals2016matching,Koch2015SiameseNN} proposed a method to learn from one training example per category, known as one-shot learning. Learning without any training example, zero-shot learning, is studied by \cite{lampert2014attribute, lei2015predicting}.

\section{Our Approach}

\begin{figure*}[t]
\begin{center}
  \includegraphics[height=0.3\textwidth]{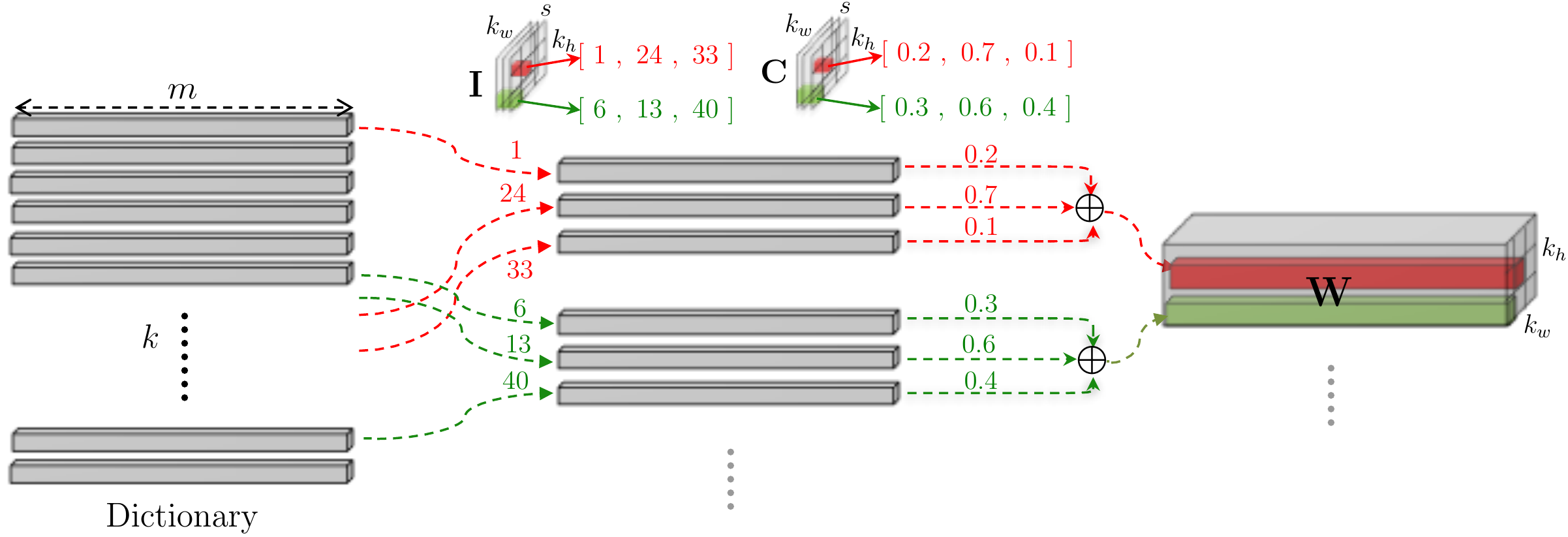}
\end{center}
\caption{This figure demonstrates the procedure for constructing a weight filter in LCNN. A vector in the weight filter (the long colorful cube in the gray tensor $\bW$) is formed by a linear combination of few vectors, which are looked up from the dictionary $\bD$. Lookup indices and their coefficients are stored in tensors $\bI$ and $\bC$.}
\label{fig:dicconv}
\end{figure*} 

\noindent \textbf{Overview:} In a CNN, each convolutional layer consists of $n$ cubic weight filters of size $m \times k_w \times k_h$, where $m$ and $n$ are the number of input and output channels, respectively, and $k_w$ and $k_h$ are the width and the height of the filter. Therefore, the weights in each convolutional layer is composed of $nk_wk_h$ vectors of length $m$. These vectors are shown to have redundant information\cite{denil2013predicting}. To avoid this redundancy, we build a relatively small set of vectors for each layer, to which we refer as dictionary, and enforce each vector in the weight filter to be a linear combination of a few elements from this set. Figure \ref{fig:dicconv} shows an overview of our model. The gray matrix at the left of the figure is the dictionary. The dashed lines show how we lookup a few vectors from the dictionary and linearly combine them to build up a weight filter. Using this structure, we devise a fast inference algorithm for CNNs. We then show that the dictionaries provide a strong prior on the visual data and enables us to learn from few examples. Finally, we show that the dictionaries can be transferred across different network architectures. This allows us to speedup the training of a deep network by transferring the dictionaries from a shallower model.



\subsection{LCNN}
\label{sec:modeldef}

\noindent A convolutional layer in a CNN consists of four parts: \begin{inparaenum}[1)] \item the input tensor $\bX \in \real^{m\times w\times h}$; where $m$, $w$ and $h$ are the number of input channels, the width and the height, respectively, \item a set of $n$ weight filters, where each filter is a tensor $\bW \in \real^{m\times k_w \times k_h}$, where $k_w$ and $k_h$ are the width and the height of the filter, \item a scalar bias term $b \in \real$ for each filter, and \item the output tensor $\bY \in \real^{n\times w' \times h'}$; where each channel $\bY_{[i,:,:]} \in \real^{w'\times h'}$ is computed by $\bW \ast \bX + b$\end{inparaenum}. Here $\ast$ denotes the discrete convolution operation\footnote{The (:) notation is borrowed from NumPy for selecting all entries in a dimension.}.


For each layer, we define a matrix $\bD \in \real^{k\times m}$ as the shared dictionary of vectors. This is illustrated in figure~\ref{fig:dicconv}, on the left side. This matrix contains $k$ row vectors of length $m$. The size of the dictionary, $k$, might vary for different layers of the network, but it should always be smaller than $nk_wk_h$, the total number of vectors in all weight filters of a layer. Along with the dictionary $\bD$, we have a tensor for lookup indices $\bI \in \natural_{\le k}^{~~s \times k_w\times k_h}$, and a tensor for lookup coefficients $\bC \in \real^{s \times k_w\times k_h}$ for each layer. For a pair $(r, c)$, $\bI_{[:,r,c]}$ is a vector of length $s$ whose entries are indices of the rows of the dictionary, which form the linear components of $\bW_{[:,r,c]}$. The entries of the vector $\bC_{[:,r,c]}$ specify the linear coefficients with which the components should be combined to make $\bW_{[:,r,c]}$ (illustrated by a long colorful cube inside the gray cub in Figure~\ref{fig:dicconv}-right). We set $s$, the number of components in a weight filter vector, to be a small number.  The weight tensor can be constructed as follows:
\begin{equation}
\label{eq:lcnnrep}
    \begin{array}{ll@{}ll}
        \bW_{[:,r,c]} = \sum\limits_{t=1}^{s} \bC_{[t,r,c]}\cdot\bD_{[\bI_{[t,r,c]}, :]} && \forall r,c
    \end{array}
\end{equation}

This procedure is illustrated in Figure~\ref{fig:dicconv}. In LCNN, instead of storing the weight tensors $\bW$ for convolutional layers, we store $\bD$, $\bI$ and $\bC$, the building blocks of the weight tensors. As a result, we can reduce the number of parameters in a convolutional layer by reducing $k$, the dictionary size, and $s$, the number of components in the linear combinations. In the next section, we will discuss how LCNN uses this representation to speedup the inference.


\begin{figure*}[t]
\begin{center}
  \includegraphics[scale=0.7]{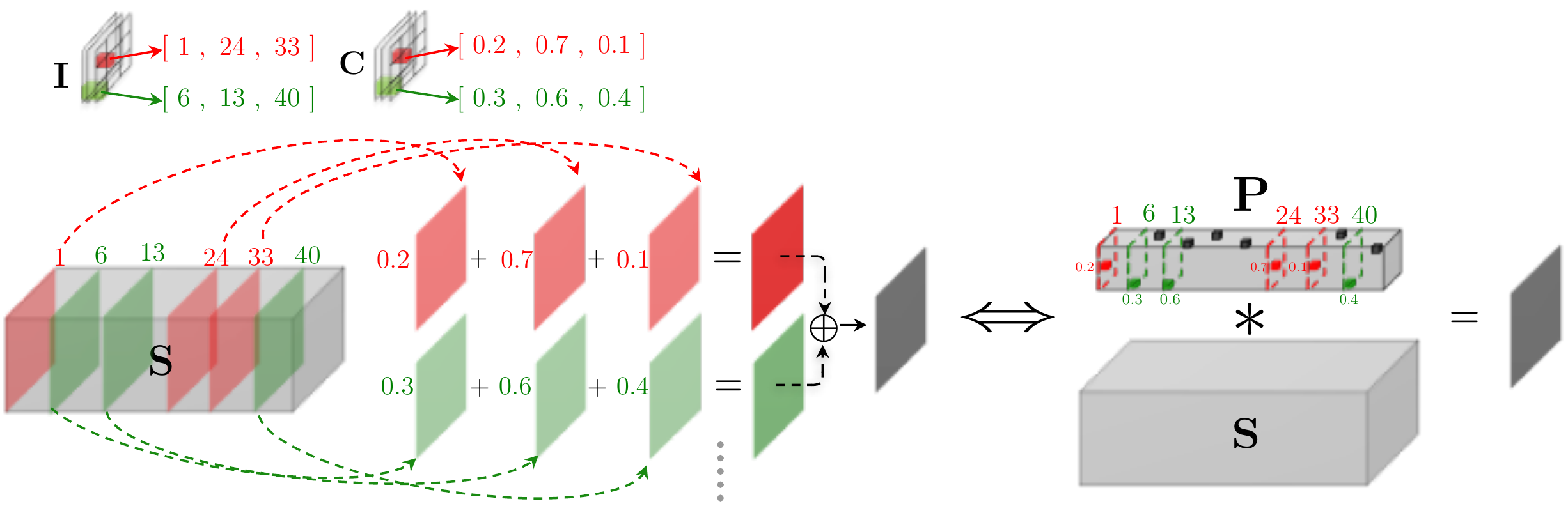}
\end{center}
\caption{$\bS$ is the output of convolving the dictionary with the input tensor. \textbf{The left side} of this figure illustrates the inference time forward pass. The convolution between the input and a weight filter is carried out by lookups over the channels of $\bS$ and a few linear combinations. Direct learning of tensors $\bI$ and $\bC$ reduces to an intractable discrete optimization. \textbf{The right side} of this figure shows an equivalent computation for training based on sparse convolutions. Parameters $\bP$ can be trained using SGD. The tiny cubes in $\bP$ denote the non-zero entries.}   
\label{fig:lookupconv}
\end{figure*}

\subsubsection{Fast Convolution using a Shared Dictionary}
\label{sec:modelspeed}
\noindent A forward pass in a convolutional layer consists of $n$ convolutions between the input $\bX$ and each of the weight filters $\bW$. We can write a convolution between an $m\times k_w\times k_h$ weight filter and the input $\bX$ as a sum of $k_wk_h$ separate $(1\times 1)$-convolutions:
\begin{equation}
\label{eq:regconv}
    \begin{array}{ll@{}ll}
        \bX\ast\bW = \sum\limits_{r,c}^{k_h, k_w} \text{shift}_{r,c}(\bX\ast\bW_{[:,r,c]}) && 
    \end{array}
\end{equation}

\noindent , where $\text{shift}_{r,c}$ is the matrix shift function along rows and columns with zero padding relative to the filter size. Now we use the LCNN representation of weights (equation \ref{eq:lcnnrep}) to rewrite each $1\times1$ convolution:
\begin{equation}
\label{eq:convrewrite}
    \begin{array}{rcl}
        \bX\ast\bW & = & \sum\limits_{r,c} \text{shift}_{r,c}(\bX\ast(\sum\limits_{t=1}^{s} \bC_{[t,r,c]}\cdot\bD_{[\bI_{[t,r,c]}, :]}))\\
                             & = & \sum\limits_{r,c} \text{shift}_{r,c}(\sum\limits_{t=1}^{s} \bC_{[t,r,c]}(\bX\ast\bD_{[\bI_{[t,r,c]}, :]}))
    \end{array}
\end{equation}

Equation \ref{eq:convrewrite} suggests that instead of reconstructing the weight tensor $\bW$ and convolving with the input, we can convolve the input with all of the dictionary vectors, and then compute the output according to $\bI$ and $\bC$. Since the dictionary ${\bD}$ is shared among all weight filters in a layer, we can pre-compute the convolution between the input tensor $\bX$ and all the dictionary vectors.
Let $\bS \in \real^{k\times w \times h}$ be the output of convolving the input $\bX$ with all of the dictionary vectors $\bD$, i.e.,
\begin{equation}
\label{eq:precomp}
    \begin{array}{ll@{}ll}
        \bS_{[i,:,:]} = \bX\ast\bD_{[i,:]} && \forall 1\le i \le k
    \end{array}
\end{equation}
Once the values of $\bS$ are computed, we can reconstruct the output of convolution by \textit{lookups} over the channels of $\bS$ according to $\bI$, then \textit{scale} them by the values in $\bC$:
\begin{equation}
\label{eq:efficientconv}
    \begin{array}{ll@{}ll}
        \bX\ast\bW = \sum\limits_{r,c}^{k_h, k_w} \text{shift}_{r,c}(\sum\limits_{t=1}^{s} \bC_{[t,r,c]}\bS_{[\bI_{[t,r,c]}, :, :]})
    \end{array}
\end{equation}

This is shown in Figure~\ref{fig:lookupconv} (left). Reducing the size of the dictionary $k$ lowers the cost of computing $\bS$ and makes the forward pass faster. Since $\bS$ is computed by a dense matrix multiplication, we are still able to use OpenBlas~\cite{openblas} for fast matrix multiplication. In addition, by pushing the value of $s$ to be small, we can reduce the number of lookups and floating point operations.


\subsubsection{Training LCNN}
\label{sec:modeltraining}
\noindent So far we have discussed how LCNN represents a weight filter by linear combinations of a subset of elements in a shared dictionary. We have also shown that how LCNN performs convolutions efficiently in two stages: \begin{inparaenum}[1-] \item \textit{Small convolutions}: convolving the input with a set of $1\times1$ filters (equation \ref{eq:precomp}). \item \textit{Lookup and scale}: few lookups over the channels of a tensor followed by a linear combination (equation \ref{eq:efficientconv}) \end{inparaenum}. Now, we explain how one can jointly train the dictionary and the lookup parameters, $\bI$ and $\bC$. Direct training of the proposed lookup based convolution leads to a combinatorial optimization problem, where we need to find the optimal values for the integer tensor $\bI$.
To get around this, we reformulate the lookup and scale stage (equation \ref{eq:efficientconv}) using a standard convolution with sparsity constraints.

Let $\bT \in \real^{k\times k_w\times k_h}$ be a one hot tensor, where $\bT_{[t,r,c]} = 1$ and all other entries are zero. It is easy to observe that convolving the tensor $\bS$ with $\bT$ will result in $\text{shift}_{r,c}(\bS_{[t,:,:]})$. We use this observation to convert the lookup and scale stage (equation~\ref{eq:efficientconv}) to a standard convolution. Lookups and scales can be expressed by a convolution between the tensor $\bS$ and a sparse tensor $\bP$, where $\bP \in \real^{ k \times w \times h}$, and $\bP_{[:,r,c]}$ is a $s$-sparse vector (\textit{i.e.} it has only $s$ non-zero entries) for all spatial positions $(r,c)$. Positions of the non-zero entries in $\bP$ are determined by the index tensor $\bI$ and their values are determined by the coefficient tensor $\bC$. Formally, tensor $\bP$ can be expressed by $\bI$ and $\bC$:
\begin{eqnarray}
\bP_{j,r,c} = 
\begin{cases}
    \bC_{t,r,c},& \exists t: \bI_{t,r,c}=j\\
    0,              & \text{otherwise}
\end{cases}
\end{eqnarray}

Note that this conversion is reversible, \ie we can create $\bI$ and $\bC$ from  the position and the values of the non-zero entries in $\bP$. With this conversion, the lookup and scale stage (equation \ref{eq:efficientconv}) becomes:
\begin{eqnarray}
\label{eq:spconv}
\begin{aligned}
  \sum_{rc} \text{shift}_{(r,c)}(\sum_{t=1}^s \bC_{[t,r,c]}\bS_{[\bI_{[t,r,c]},:,:]}) = \bS \ast \bP
\end{aligned}
\end{eqnarray}
This is illustrated in Figure~\ref{fig:lookupconv}-right. Now, instead of directly training $\bI$ and $\bC$, we can train the tensor $\bP$ with $\ell_0$-norm constraints ($\norm{\bP_{[:,r,c]}}_{\ell_0} = s$) and then construct $\bI$ and $\bC$ from $\bP$. However, $\ell_0$-norm is a non-continuous function with zero gradients everywhere. As a workaround, we relax it to $\ell_1$-norm. At each iteration of training, to enforce the sparsity constraint for $\bP_{[:,r,c]}$, we sort all the entries by their absolute values and keep the top $s$ entries and zero out the rest. During training, in addition to the classification loss $L$ we also minimize $\sum\limits_{[r,c]} \norm{\bP_{[:,r,c]}}_{\ell_1} = \norm{\bP}_{\ell_1}$, by adding a term $\lambda\norm{\bP}_{\ell_1}$ to the loss function. The gradient with respect to the values in $\bP$ is computed by:
\begin{eqnarray}
\label{eq:grad}
\begin{aligned}
  \frac{\partial (L+\lambda \left\| \bP \right\|_{\ell_1})}{\partial \bP}= \frac{\partial L}{\partial \bP}+\lambda ~\text{sign}(\bP)
\end{aligned}
\end{eqnarray}
where $\frac{\partial L}{\partial \bP}$ is the gradient that is computed through a standard back-propagation. $\lambda$ is a hyperparameter that adjusts the trade-off between the CNN loss function and the $\ell_1$ regularizer.
We can also allow $s$, the sparsity factor, to be different at each spatial position $(r,c)$, and be determined automatically at training time. This can be achieved by applying a threshold function,
\begin{eqnarray}
\label{eq:threshold}
\begin{aligned}
  \delta(x) = \begin{cases} x, & \left|x\right|>\epsilon \\ 0, & \text{otherwise} \end{cases}
\end{aligned}
\end{eqnarray}
 over the values in $\bP$ during training. We also backpropagate through this threshold function to compute the gradients with respect to $\bP$. The derivative of the threshold function is $1$ everywhere except at $|x| < \epsilon$, which is $0$. Hence, if any of the entries of $\bP$ becomes $0$ at some iteration, they stay $0$ forever. Using the threshold function, we let each vector to be a combination of arbitrary vectors. At the end of the training, the sparsity parameter $s$ at each spatial position $(r,c)$ is determined by the number of non-zero values in $\bP{[:,r,c]}$.

Although the focus of our work is to speedup convolutional layers where most of the computations are, our lookup based convolution model can also be applied on fully connected (FC) layers. An FC layer that goes from $m$ inputs to $n$ outputs can be viewed as a convolutional layer with input tensor $m\times1\times1$ and $n$ weight filters, each of size $m\times1\times1$. We take the same approach to speedup fully connected layers.

After training, we convert $\bP$ to the indices and the coefficients tensors $\bI$ and $\bC$ for each layer. At test time, we follow equation~\ref{eq:efficientconv} to efficiently compute the output of each convolutional layer.



\subsection{Few-shot learning}
The shared dictionary in LCNN allows a neural network to learn from very few training examples on novel categories, which is known as few-shot learning\cite{hariharan2016low}. A good model for few-shot learning should have two properties: \begin{inparaenum}[a)]\item strong priors on the data, and \item few trainable parameters\end{inparaenum}. LCNN has both of these properties. An LCNN trained on a large dataset of images (e.g.\ ImageNet~\cite{imagenet_cvpr09}) will have a rich dictionary $\bD$ at each convolutional layer.  This dictionary provides a powerful prior on visual data. At the time of fine-tuning for a new set of categories with few training examples, we only update the coefficients in $\bC$. This reduces the number of trainable parameters significantly.    

In a standard CNN, to use a pre-trained network to classify a set of novel categories, we need to reinitialize the classification layer randomly. This introduces a large number of parameters, on which we don't have any prior, and they should be trained solely by a few examples. LCNN, in contrast, can use the dictionary of the classification layer of the pre-trained model, and therefore only needs to learn $\bI$ and $\bC$ from scratch, which form a much smaller set of parameters. Furthermore, for all other layers, we only fine-tune the coefficients $\bC$, \ie only update the non-zero entries of $\bP$. Note that the dictionary $\bD$ is fixed across all layers during the training with few examples.

\subsection{Few-iteration learning}
\label{sec:few-iteration}
Training very deep neural networks are computationally expensive and require hundreds of thousands of iterations. This is mainly due to the complexity of these models. In order to constrain the complexity, we should limit the number of learnable parameters in the network. LCNN has a suitable setting that allows us to limit the number of learnable parameters without changing the architecture. This can be done by transferring the shared dictionaries $\bD$ from a shallower network to a deeper one.

Not only we can share a dictionary $\bD$ across layers, but we can also share it across different network architectures of different depths. A dictionary $\bD \in \real^{m\times k}$ can be used in any convolutional layer with input channel size $m$ in any CNN architecture. For example, we can train our dictionaries on a shallow CNN and reuse in a deeper CNN with the same channel size. On the deeper CNN we only need to train the indices and coefficients tensors $\bI$ and $\bC$.

\section{Experiments}
We evaluate the accuracy and the efficiency of LCNN under different settings. We first evaluate the accuracy and speedup of our model for the task of object classification, evaluated on the standard image classification challenge of ImageNet, ILSRVC2012~\cite{imagenet_cvpr09}.
We then evaluate the accuracy of our model under few-shot setting. We show that given a set of novel categories with as small as $1$ training example per category, our model is able to learn a classifier that is both faster and more accurate than the CNN baseline. 
Finally we show that the dictionaries trained in LCNN are generalizable and can be transferred to other networks. This leads to a higher accuracy in small number of iterations compared to standard CNN.



\begin{table}
\centering
\small
\begin{tabular}{|l | c | c | c |}
 \hline
 & \multicolumn{3}{|c|}{\textbf{AlexNet}}\\
 \hline
 \textbf{Model} & \textbf{speedup} & \textbf{top-1} & \textbf{top-5} \\
 \Xhline{4\arrayrulewidth}
 CNN & $1.0\times$ & $56.6$ & $80.2$ \\
 \hline
 Wen \etal \cite{wen2016learning} & $3.1\times$\tablefootnote{They have not reported the overall speedup on AlexNet, but only per layer speedup. $3.1\times$ is the weighted average of their per layer speedups.} & $\mathbf{55.4}$ & N/A\\
 \hline
 XNOR-Net\cite{Rastegari2016XNORNetIC} & $8.0\times$\tablefootnote{XNOR-Net gets $32\times$ layer-wise speedup on a $32$ bit machine. However, since they haven't binarized the first and the last layer (which has $9.64\%$ of the computation), their overall speedup is $8.0\times$.} & $44.2$ & $69.2$\\
 \Xhline{3\arrayrulewidth}
 LCNN-fast & $\mathbf{37.6}\times$ & $44.3$ & $68.7$\\
 \hline
 LCNN-accurate & $3.2\times$ & $\mathbf{55.1}$ & $\mathbf{78.1}$\\
 \hline
\end{tabular}
\vspace{3mm}
\caption{Comparison of different efficient methods on AlexNet. The accuracies are classification accuracy on the validation set of ILSVRC2012.}
\label{tab:alexnet}
\end{table}

\subsection{Implementation Details}
\label{sec:impl}
We follow the common way of initializing the convolutional layers by Gaussian distributions introduced in \cite{glorot2010understanding}, including for the sparse tensor $\bP$. We set the threshold in equation \ref{eq:threshold} for each layer in such a way that we maintain the same initial sparsity across all the layers.
That is, we set the threshold of each layer to be $\epsilon = c\cdot\sigma$, where $c$ is constant across layers and $\sigma$ is the standard deviation of Gaussian initializer for that layer. We use $c=0.01$ for AlexNet and $c=0.001$ for ResNet. Similarly, to maintain the same level of sparsity across layers we need a $\lambda$ (equation \ref{eq:grad}) that is proportional to the standard deviation of the Gaussian initializers. We use $\lambda = \lambda'\epsilon$, where $\lambda'$ is constant across layers and $\epsilon$ is the threshold value for that layer. We try $\lambda' \in \{0.1, 0.2, 0.3\}$ for both AlexNet and ResNet to get different sparsities in $\bP$.

The dictionary size $k$, the regularizer coefficient $\lambda$, and threshold value $\epsilon$ are the three important hyperparameters for gaining speedup. The larger the dictionary is, the more accurate (but slower) the model becomes. The size of the the dictionary for the first layer does not need to be very large as it's representing a $3$-dimensional space. We observed that for the first layer, a dictionary size as small as $3$ vectors is sufficient for both AlexNet and ResNet. In contrast, fully connected layers of AlexNet are of higher dimensionality and a relatively large dictionary is needed to cover the input space. We found dictionary sizes $512$ and $1024$ to be proper for fully connected layers. In AlexNet we use the same dictionary size across other layers, which we vary from $100$ to $500$ for different experiments. In ResNet, aside from the very first layer, all the other convolutional layers are grouped into $4$ types of ResNet blocks. The dimensionality of input is equal between same ResNet block types, and is doubled for consecutive different block types. In a similar way we set the dictionary size for different ResNet blocks: equal between the same block types, and doubles for different consecutive block types. We vary the dictionary size of the first block from $16$ to $128$ in different experiments.

\begin{table}
\centering
\small
\begin{tabular}{|l | c | c | c |}
 \hline
 & \multicolumn{3}{|c|}{\textbf{ResNet-18}}\\
 \hline
 \textbf{Model} & \textbf{speedup} & \textbf{top-1} & \textbf{top-5} \\
 \Xhline{4\arrayrulewidth}
 CNN & $1.0\times$ & $69.3$ & $90.0$ \\
 \hline
 XNOR-Net\cite{Rastegari2016XNORNetIC} & $10.6\times$ & $51.2$ & $73.2$\\
 \Xhline{3\arrayrulewidth}
 LCNN-fast & $\mathbf{29.2}\times$ & $51.8$ & $76.8$\\
 \hline
 LCNN-accurate & $5\times$ & $\mathbf{62.2}$ & $\mathbf{84.6}$\\
 \hline
\end{tabular}
\vspace{3mm}
\caption{Comparison of LCNN and XNOR-Net on ResNet-18. The accuracies are classification accuracy on the validation set of ILSVRC2012.}
\label{tab:resnet18}
\end{table}

\begin{figure}[b]
\begin{center}
  \includegraphics[width=0.4\textwidth]{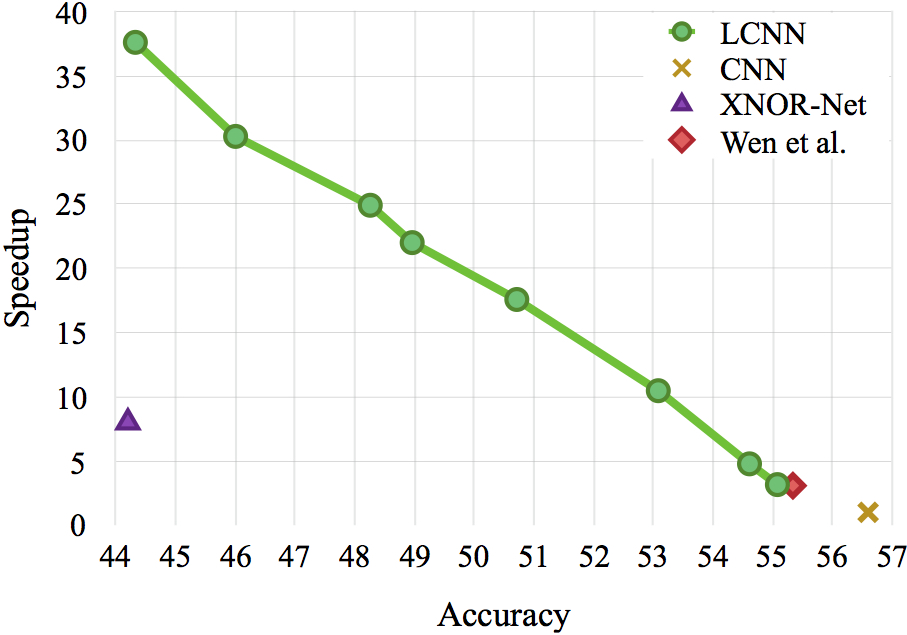}
\end{center}
\caption{Accuracy vs.\ speedup. By tuning the dictionary size, LCNN achieves a spectrum of speedups.}
\label{fig:alexnetplot}
\end{figure}

\begin{figure*}[t]
\centering
\subfigure[cats, sofas and bicycles excluded]{
  \includegraphics[width=0.4\textwidth]{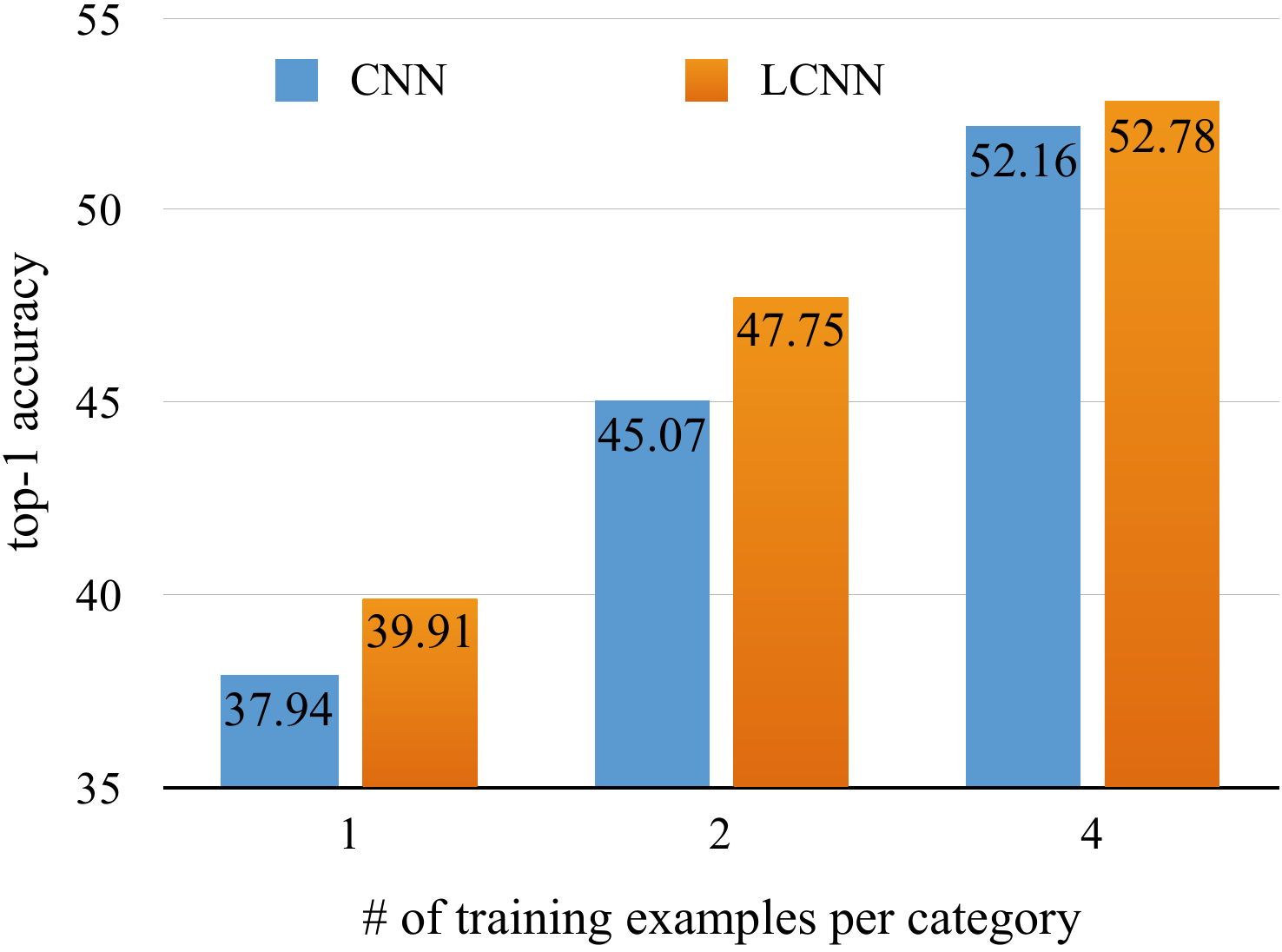}
}
\subfigure[10 random categories excluded]{
  \includegraphics[width=0.4\textwidth]{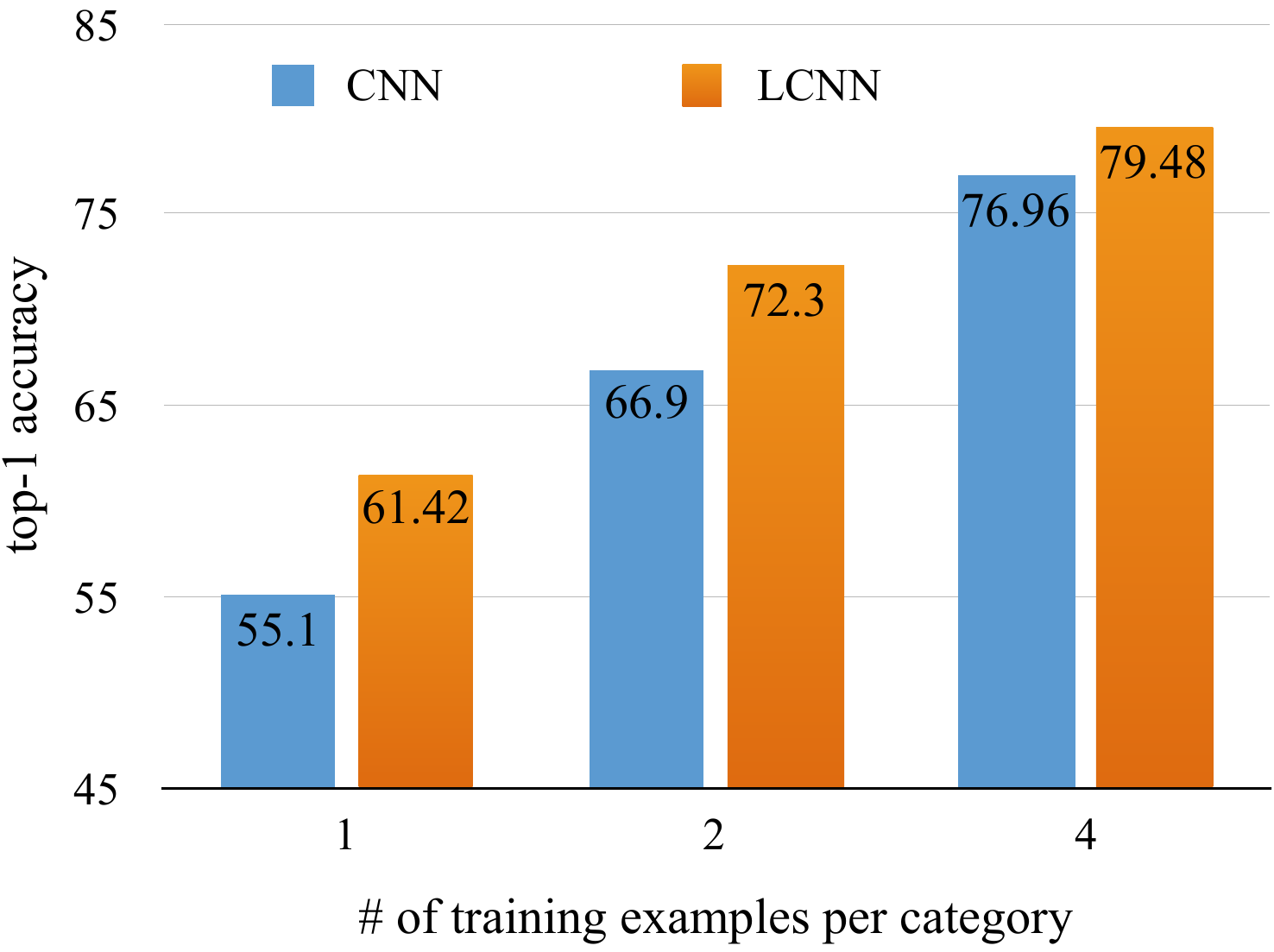}
}
\caption{Comparison between the performance of LCNN and CNN baseline on few-shot learning, for $\{1, 2, 4\}$ examples per category. In (a) all cats ($7$ categories), sofas ($1$ category) and bicycles ($2$ categories) are held out for few-shot learning. In (b), $10$ random categories are held out for few-shot learning. We repeat sampling the $10$ random categories $5$ times to avoid over-fitting to a specific sampling.}
\label{fig:fewexample}
\end{figure*}
\subsection{Image Classification}
In this section we evaluate the efficiency and the accuracy of LCNN for the task of image classification. Our proposed lookup based convolution is general and can be applied on any CNN architecture. We use AlexNet~\cite{krizhevsky2012imagenet} and ResNet~\cite{kaming2016residual} architectures in our experiments. We use ImageNet challenge ILSVRC2012~\cite{imagenet_cvpr09} to evaluate the accuracy of our model. We report standard top-$1$ and top-$5$ classification accuracy on $1$K categories of objects in natural scenes. To evaluate the efficiency, we compare the number of floating point operations as a representation for speedup. The speed and the accuracy of our model depend on two hyperparameters: \begin{inparaenum}[1)] \item $k$, the dictionary size and \item $\lambda$, which controls the sparsity of $\bP$; \ie the average number of dictionary components in the linear combination \end{inparaenum}. One can set a trade-off between the accuracy and the efficiency of LCNN by adjusting these two parameters.
We compare our model with several baselines: \begin{inparaenum}[1-]\item XNOR-Net~\cite{Rastegari2016XNORNetIC}, which reduces the precision of weights and outputs to $1$-bit, and therefore multiplications can be replaced by binary operations. In XNOR-Net, all the layers are binarized except the first and the last layer (in AlexNet, they contain $9.64\%$ of the computation). \item Wen \etal~\cite{wen2016learning}, which speeds up the convolutions by sparsifying the weight filters\end{inparaenum}.

Table~\ref{tab:alexnet} compares the top-$1$ and top-$5$ classification accuracy of LCNN with baselines on AlexNet architecture. It shows that with small enough dictionaries and sparse linear combinations, LCNN offers $37.6\times$ speedup with the accuracy of XNOR-Net. On the other hand, if we set the dictionaries to be large enough, LCNN can be as accurate as slower models like Wen \etal. In LCNN-fast, the dictionary size of the mid-layer convolutions is $30$ and for the fully connected layers is $512$. In LCNN-accurate, the mid-layer convolutions have a dictionary of size $500$ and the size of dictionary in fully connected layers is $1024$. The reguralizer constant (Section \ref{sec:impl}) $\lambda'$ for LCNN-fast and LCNN-accurate is $0.3$ and $0.1$, respectively.


Depending on the dictionary size and $\lambda'$, LCNN can achieve various speedups and accuracies. Figure~\ref{fig:alexnetplot} shows different accuracies vs.\ speedups that our model can achieve. The accuracy is computed by top-1 measure and the speedup is relative to the original CNN model. It is interesting to see that the trend is nearly linear. The best fitted line has a slope of $-3.08$, \ie  for each one percent accuracy that we sacrifice in top-1, we gain $3.08$ more speedup.


We also evaluate the performance of LCNN on ResNet-18 architecture. ResNet-18 is a compact architecture, which has $5\times$ fewer parameters in compare to AlexNet while it achieves $12.7\%$ higher top-1 accuracy. That makes it a much more challenging architecture for further compression. Yet we show that we can gain large speedups with a few points drop in the accuracy. Table~\ref{tab:resnet18} compares the accuracy of LCNN, XNOR-Net~\cite{Rastegari2016XNORNetIC}, and the original model (CNN). LCNN-fast is getting the same accuracy as XNOR-Net while getting a much larger speedup. Moreover, LCNN-accurate is getting a much higher accuracy yet maintaining a relatively large speedup. LCNN-fast has dictionaries of size $16$, $32$, $64$, and $128$ for different block types. LCNN-accuracte has larger dictionaries: $128$, $256$, $512$ and $1024$ for different block types. 

\subsection{Few-shot Learning}
In this section we evaluate the performance of LCNN on the task of few-shot learning. To evaluate the performance of LCNN on this task, we split the categories of ImageNet challenge ILSVRC2012 into two sets: \begin{inparaenum}[i)]\item base categories, a set of $990$ categories which we use for pre-training, and \item novel categories, a set of $10$ categories that we use for few-shot learning.\end{inparaenum} We do experiments under $1$, $2$, and $4$ samples per category. We take two strategies for splitting the categories. One is random splitting, where we randomly split the dataset into $990$ and $10$ categories. We repeat the random splitting $5$ times and report the average over all. The other strategy is to hold out all cats ($7$ categories), bicycles ($2$ categories) and sofa ($1$ category) for few-shot learning, and use the other $990$ categories for pre-training. With this strategy we make sure that base and novel categories do not share similar objects, like different breeds of cats. For each split, we repeat the random sampling of $1$, $2$, and $4$ training images per category $20$ times, and get the average over all. \fixme{Repeating the random sampling of the few examples is crucial for any few-shot learning experiment, since a model can easily overfit to a specific sampling of images.}





We compare the performance of CNN and LCNN on few-shot learning in Figure~\ref{fig:fewexample}. We first train an original AlexNet and an LCNN AlexNet on all training images of base categories ($990$ categories, $1000$ images per category). We then replace the $990$-way classification layer with a randomly initialized $10$-way linear classifier. In CNN, this produces $10\times4096$ randomly initialized weights, on which we don't have any prior. These parameters need to be trained merely from the few examples. In LCNN, however, we transfer the dictionary trained in the $990$-way classification layer to the new $10$-way classifier. This reduces the number of randomly initialized parameters by at least a factor of $4$. We use AlexNet LCNN-accurate model (same as the one in Table~\ref{tab:alexnet}) for few-shot learning. At the time of fine-tuning for few-shot categories, we keep the dictionaries in all layers fixed and only fine-tune the sparse $\bP$ tensor. This reduces the total number of parameters that need to be fine-tuned by a factor of $14\times$. We use different learning rates $\eta$ and $\eta'$ for the randomly initialized classification layer (which needs to be fully trained) and the previous pre-trained layers (which only need to be fine-tuned). We tried $\eta' = \eta$, $\eta' = \frac{\eta}{10}$, $\eta' = \frac{\eta}{100}$ and $\eta' = 0$ for both CNN and LCNN, then picked the best for each configuration.

Figure~\ref{fig:fewexample} shows the top-1 accuracies of our model and the baseline in the two splitting strategies of our few-shot learning experiment. In Figure~\ref{fig:fewexample}~(a) we are holding out all cat, sofa, and bicycle categories ($10$ categories in total) for few-shot learning. LCNN is beating the baseline consistently in $\{1, 2, 4\}$ examples per category. Figure~\ref{fig:fewexample}~(b) shows the comparison in the random splitting strategy. We repeat randomly splitting the categories into $990$ and $10$ categories $5$ times, and report the average over all. Here LCNN gets a larger improvement in the top-1 accuracy compared to the baseline for $\{1, 2, 4\}$ images per category.


\begin{figure}[t]
\begin{center}
  \includegraphics[width=0.48\textwidth]{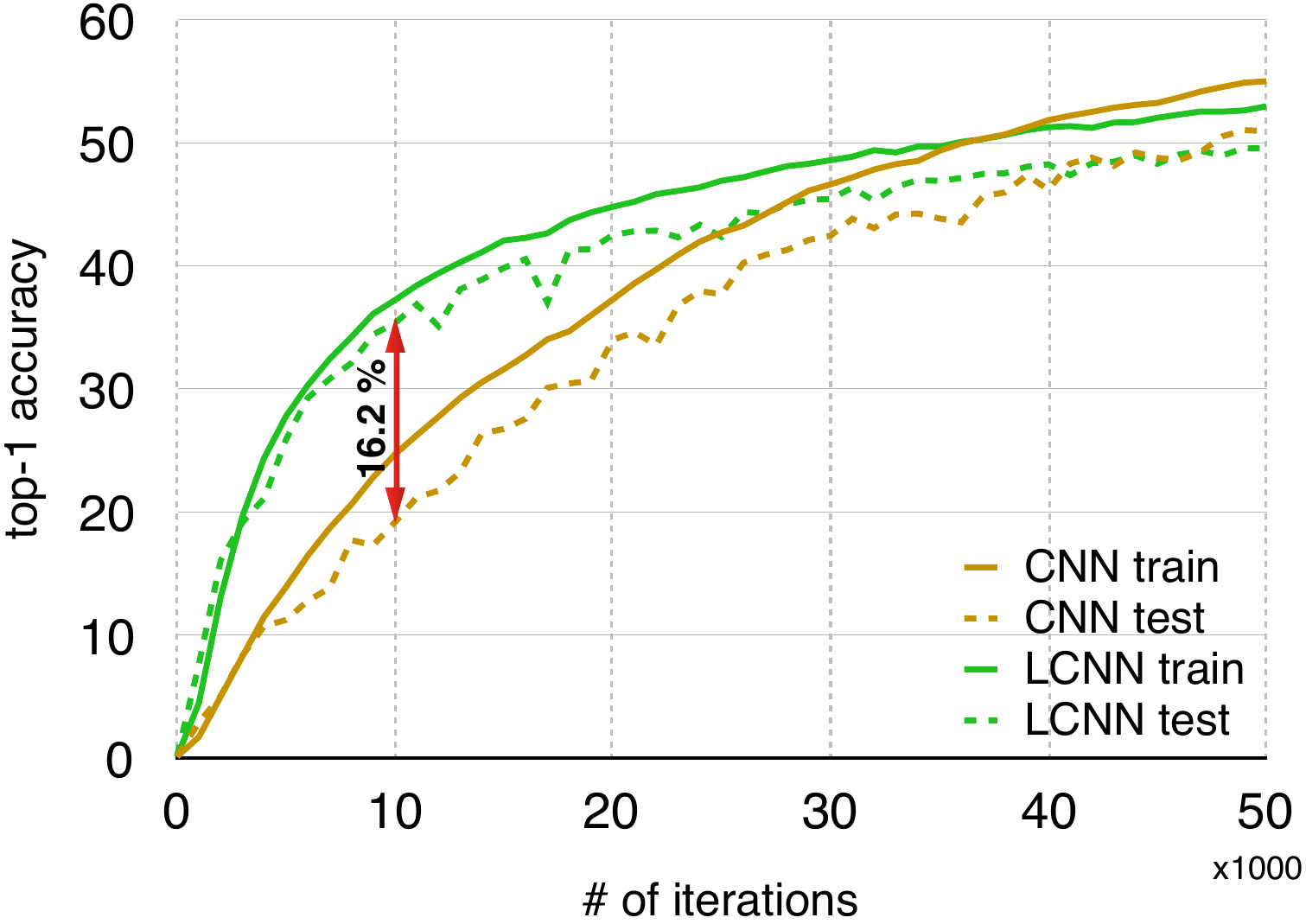}
\end{center}
\vspace{-0.5cm}
\caption{\small LCNN can obtain higher accuracy on few iterations by transferring the dictionaries $\bD$ from a shallower architecture. This figure illustrates the learning curves on top-1 accuracy for both LCNN and standard CNN. The accuracy of LCNN is $16.2\%$ higher than CNN at iteration $10K$. }
\vspace{-0.2cm}
\label{fig:few-iterations}
\end{figure}

\subsection{Few-iteration Learning}
In section~\ref{sec:few-iteration} we discussed that the dictionaries in LCNN can be transferred from a shallower network to a deeper one. As a result, one can train fewer parameters--only $\bI$ and $\bC$--in the deeper network with few iterations obtaining a higher test accuracy compared to a standard CNN. In this experiment we train a ResNet with $1$ block of each type, 10 layers total. We then transfer the dictionaries of each layer to its corresponding layer of ResNet-18 (with 18 layers). After transfer, we keep the dictionaries fixed. We show that we get higher accuracy in small number of iterations compared to standard CNN. Figure~\ref{fig:few-iterations} illustrates the learning curves on top-1 accuracy for both LCNN and standard CNN. The test accuracy of LCNN is $16.2\%$ higher than CNN at iteration $10K$. The solid lines denote the training accuracy and the dashed lines denote the test accuracy.


\section{Conclusion}
 With recent advancements in virtual reality,  augmented reality, and smart wearable devices, the need for getting the state of the art deep learning algorithms onto these resource constrained compute platforms increases. Porting state of the art deep learning algorithms to resource constrained compute platforms is extremely challenging.  We introduce LCNN, a lookup-based convolutional neural network that encodes convolutions by few lookups to a dictionary that is trained to cover the space of weights in CNNs.  Training LCNN involves jointly learning a dictionary and a small set of linear combinations. The size of the dictionary naturally traces a spectrum of trade-offs between efficiency and accuracy. 
 
 LCCN enables efficient inference; our experimental results on ImageNet challenge show that LCNN can offer $3.2\times$ speedup while achieving $55.1\%$ top-1 accuracy using AlexNet architecture. Our fastest LCNN offers $37.6\times$ speed up over AlexNet while maintaining $44.3\%$ top-1 accuracy. LCNN not only offers dramatic speed ups at inference, but it also enables efficient training. On-device training of deep learning methods requires algorithms that can handle few-shot and few-iteration constrains. LCNN can simply deal with these problems because our dictionaries are architecture agnostic and transferable across layers and architectures, enabling us to only learn few linear combination weights. Our future work involves exploring low-precision dictionaries as well as compact data structures for the dictionaries.


\noindent {\bf Acknowledgments:} This work is in part supported by ONR N00014-13-1-0720, NSF IIS-1338054,  NSF-1652052, NRI-1637479, Allen Distinguished Investigator Award, and the Allen Institute for Artificial Intelligence.

{\small
\bibliographystyle{ieee}
\bibliography{bib}
}

\end{document}